# Embed System for Robotic Arm with 3 Degree of Freedom Controller using Computational Vision on Real-Time

Luiz Cortinhas[1], Patrick Monteiro[1], Amir Zahlan[1], Gabriel Vianna[1] and Marcio Moscoso[2]

[1]Instituto de Ensinos Superiores da Amazonia, Belém, Pará
iesam@iesam.com.br
[2]Instituto Federal do Pará, Belém, Pará
ifpa@ifpa.edu.br

## ABSTRACT

*This Paper deals with robotic arm embed controller system, with distributed system based on protocol communication between one server supporting multiple points and mobile applications trough sockets .The proposed system utilizes  hand with glove gesture in three-dimensional recognition using fuzzy implementation to set x,y,z coordinates. This approach present all implementation over: two raspberry PI arm based computer running client program, x64 PC running server program, and one robot arm controlled by ATmega328p based board.*

## 1. INTRODUCTION

Gesture Recognition is an important, yet difficult task on arm-based embeds systems [1].
It is a versatile and intuitive way to approach the more natural form to human-machine interaction just need glove with five light emissor diode at fingertips, tracking the movements over filtered images sequences captured by webcam and recovering data to 3D structure on real time. At same time arm-based system is a low-cost and newest trend to approach mobile world, these reasons make Raspberry PI the best choice. This presented is designed to JavaSE version 7 solution because this code language is perfect to minimize code creations for different architectures[4] and compiled programs owing to virtual machine developed to all specified devices architecture: X86, X64 and ARM, all these running a generic Linux SO.

## 2. THE GLOVE

To avoid complex implementations of Image Processing and obtain major precision was built a simple glove fig.1 on fingertips located LEDs with a wavelength 850 nm (infra-red) and 5mm of diameter. The glove has a 3,3v coin battery that energizes the five leds through different five 220 ohm resistors. This implementation is very cheap and capable to make more smooth detected movements.

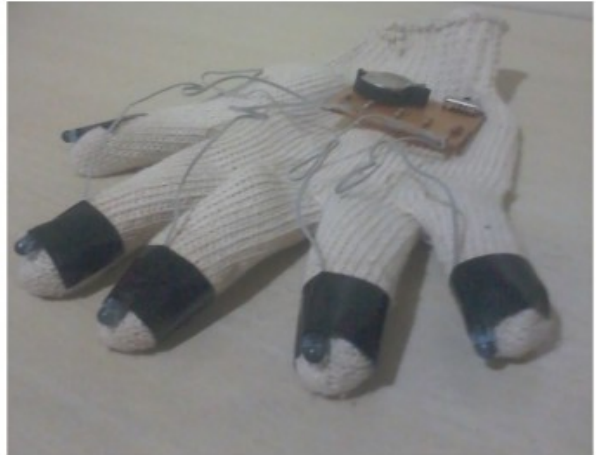
Figure 1. The Glove

### 3. The Hardware

Except by the glove described above, there are two hardwares: Minimum Computer Requirements and Robot Arms.

  A.  Minimum Computer Requirements

Considering the propose to support cheap computers and embed systems, the minimum requirements is based on raspberry PI requirements: processor at 700 Mhz, 384 Mb de RAM, 128 Mb for Video Memory and at least 1Mbps connection.

  B.  Robotic Arm

The robotic arm on fig.8 is named Mark II build in aluminium on thickness of 1mm in four body parts:

  -  Base: Built on box on (16 cm x 16 cm x 11 cm) format content power supply AC/DC converter with 12v at 5A and 5v at 1A outputs, 12 v is used to energize: four servo motor model MG995 – TowerPro each projected to load 15kg/cm on maximum and one, the 5 v Supply is used to energize atmega328P responsible to convert input data in degrees to coordinates movements, supporting two servo motors the base weight proximate at 3kg.
  -  Lower Arm: starting from base joint measuring 40 cm in length and 4cm in width, supports 1 servo motor on upper end, this element weight 180 g measured.
  -  Higher Arm: Starting from Lower Arm joint above described measuring 20 cm in length and 2,7cm in width, support one servo motor on upper end, weight 110 g measured.
  -  Claw Grip: For purposes especial to minimal weight and force to grip function, the claw use a little servo motor model TowerPro MicroServo 9 g – SG90 measuring 9 g and concentrate force 1kg/cm in maximum load, generate a higher grip force, measured all weight 20 g.

### 4. Computational Vision

This section is focused on motion analysis of hand with glove presented revealed that gesture can be characterized based on four different aspects: shape, motion, position and orientation. All gesture recognition approaches try to approach the problem by concentrating on one or more of above aspects according in [1], on this case will go use position and orientation. To solve problem of low hardware specification adopted: this method of recognition is easiest using just one hand whit five dots to interpretation.

The motion analysis start at first captured sequence images on second step the image processing is applied on source the High Pass algorithm developed show only higher value pixels, works scaling between black to white on black pixel equal 0 and white pixel equal 255, this method is developed to dynamical approach to solve problem of different conditions of illumination every searching minor and major pixel and set interval minimum and maximum range on scale, the High Pass algorithm filter for pixel value between 240 and 255 finally the image is converted to binary format fig.2, on next step is applied Hough Transform[2] searching for circular shape [5]-[6] obtain only position of five most approximate diameter shapes calculate center of white shapes shown each as blue circles to calculate the hand's center fig.4 generating coordinate to be displayed on red dot with yellow trace, it all is shown in fig.3,if this is a first capture this hand's center coordinates are saved and adjusted to zero on next image all processes above will happen and newest "hand's center" coordinates (x,y and z) and last "hand's center" calculate the difference on each resulting on pixel moved parsing to Fuzzy Logic, except Z, that will define the real movement to be sent to robotics arm on angular form ranged (0º-180º).

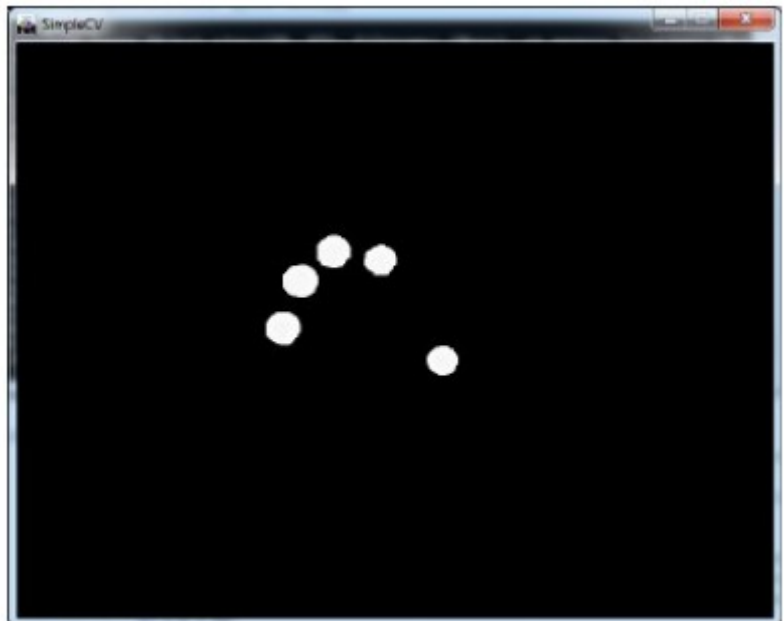

Figure 2. Result of High Pass Filter Algorithm applied on all pixels

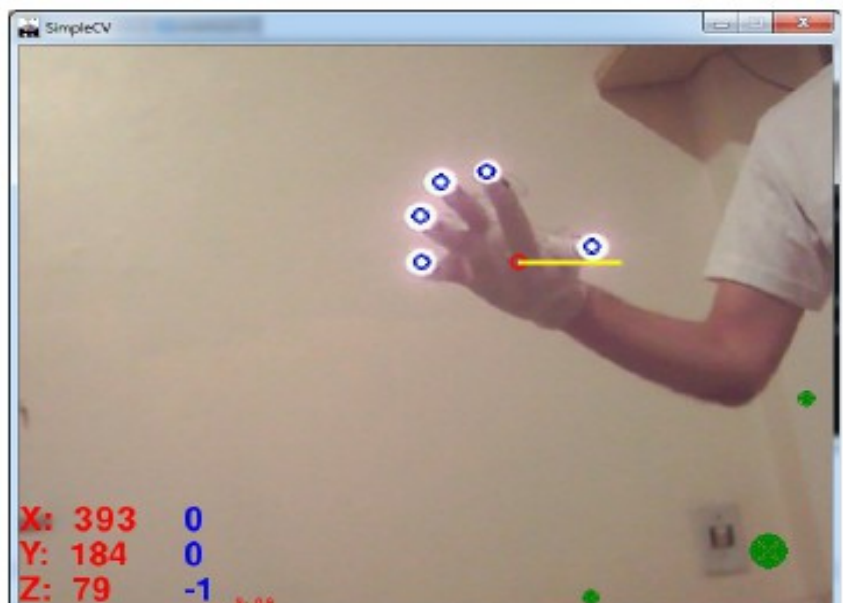

Figure 3. Result of Analysis and Image Processing

$$xCenter = \frac{\sqrt{((max(x) - min(x))^2)}}{2},$$

$$yCenter = \frac{\sqrt{((max(y) - min(y))^2)}}{2},$$

$$zCenter = (\frac{\sum x - xCenter}{5}) - (\frac{\sum y - yCenter}{5})$$

Figure 4. Formulas to calculate hand's center point.

## 5. Fuzzy Logic

The fuzzy is a multivalued logic responsible to transform not exact information on acceptable output supporting stochastic outputs based on rules, this way approximate human to machine [3] this start with the receipt of the amount of displacement and the direction of movement by the center of the hand applying the rules of fig.5 for movement on X axis and fig.6 for Y axis, each of these rules on axis x represent the number of pixels displaced and y axis the degree of relevance. For these entrances fuzzy generate one output based on result of each entrance transformed by respective rule see output rule on fig.7 to X and Y, verify this important output necessary to eliminate the problem of precise movements of the human hand. The fuzzy resultant is sent to robotic arm on format: "(x,y,z)" each on their degree information result.

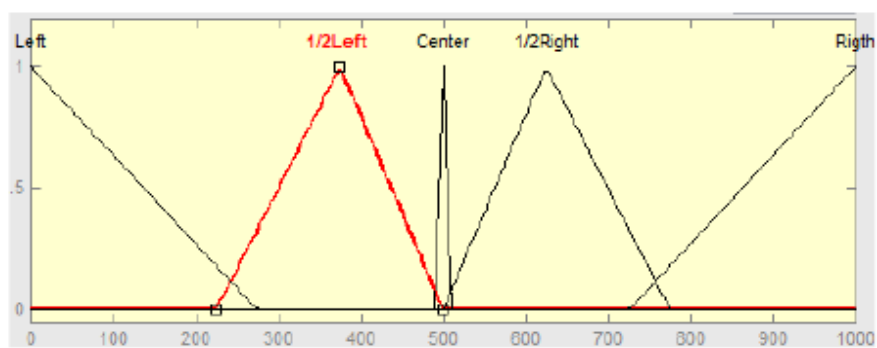
Figure 5. Fuzzy rule Input for x axis.

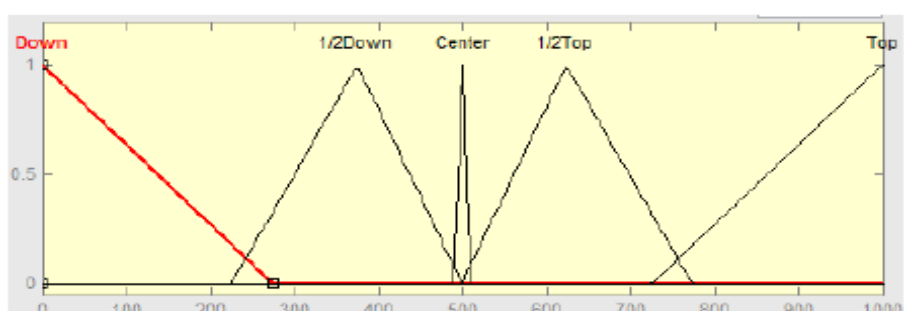
Figure 6. Fuzzy rule Input for y axis.

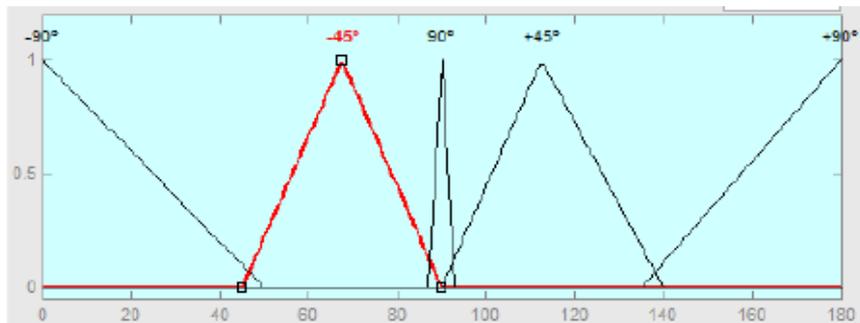
Figure 7. Fuzzy Logic rule Output on degrees

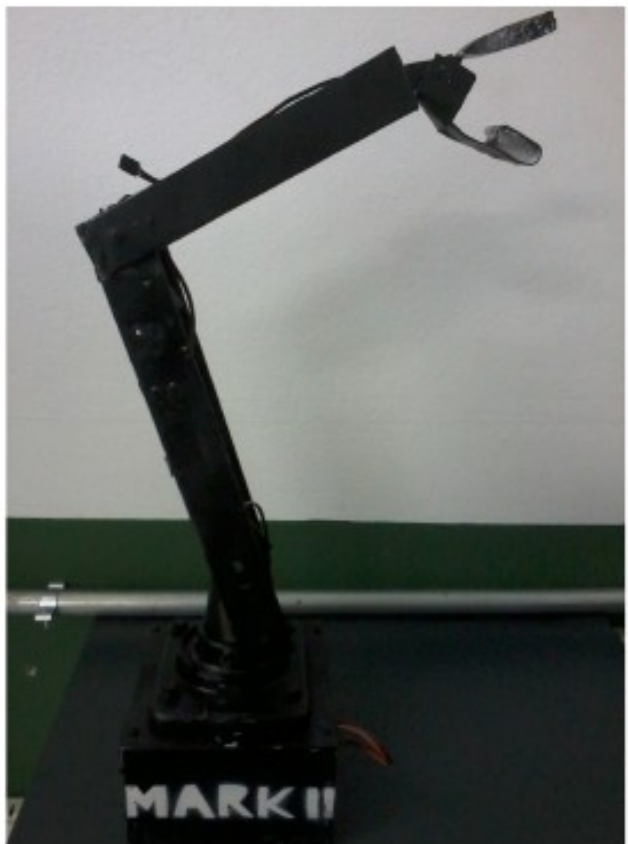
Figure 8. Robotics Arm Mark II (Stable Version)

## 3. CONCLUSIONS

This presented work all objectives are presented and built a possible embed system using processing image on acceptable scale interacting with hardware robotics arm developed to symbolize an industrial robotic arm and show that the relationship man - machine can become more human and get better results from movement with little investment, besides the mobility applied allowing almost any machine can be included in the system. The major process included on this process is presented in fig.9 below, this project reaches its goals and demonstrates how ARM architecture is being inserted how low-cost machine.

The robotic arm demonstrate stability and security same in poor conditions of hand stability, because the Fuzzy Model works on stability controller in 2D (X,Y) the Z axis is don't demonstrate instability because this is calculated using means minimizing little movements[7].

On tests Raspberry Pi with client running only receives Camera Images only applying Movement Analysis, Fuzzy and Send Data to hardware by way of Serial UART (at 9600bps), this approach reached goal with successful movements fast and accurate.

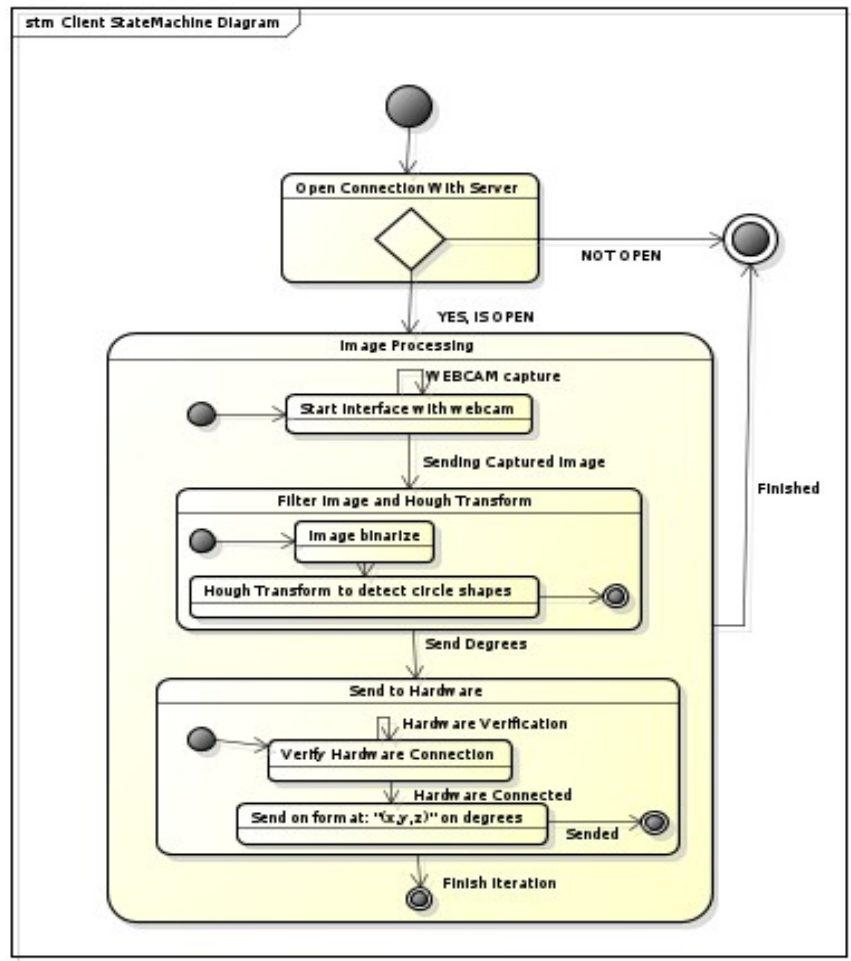

Figure 9. State Machine Diagram Overall Client Program

## ACKNOWLEDGEMENTS

Thanks everyone include my literary agent!

**Authors**

Luiz Cortinhas Ferreira Neto

Short Biography

Graduating on Computer Engineering – IESAM since 2010

Member of LINC – Laboratory of Inteligence Computational / UFPA

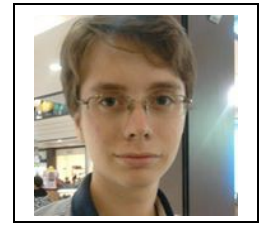

Gabriel Vianna Soares Rocha

Graduating on Computer Engineering – IESAM since 2010

Member of LINC – Laboratory of Inteligence Computational

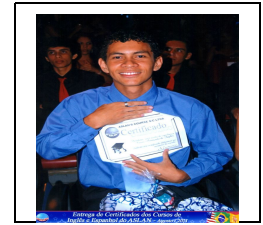

Amir Samer Zahlan

Graduating on Computer Engineering – IESAM since 2010

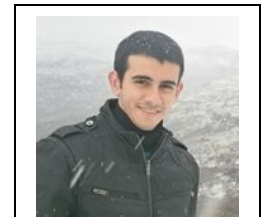

Patrick Monteiro

Graduating on Computer Engineering – IESAM since 2010

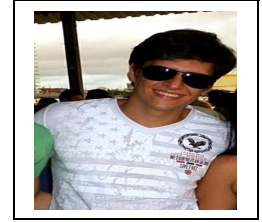

Marcio Nazareno de Araujo Moscoso

Master on Eletrical Engineer on UFPB, since 200

Graduated on Eletrical Engineer on UFPA, since 1998

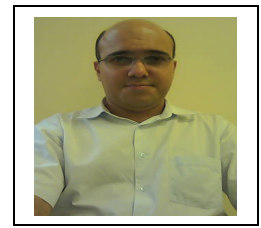